\documentclass[10pt,journal,compsoc]{IEEEtran}
\newif\ifpeerreview

\peerreviewfalse

\usepackage[nocompress]{cite}
\usepackage{url}
\usepackage{amsmath,amssymb,graphicx}

\usepackage{amsmath,amsfonts}
\usepackage{amsmath,amssymb,bm}

\usepackage{algorithmic}
\usepackage{algorithm}
\usepackage{array}
\usepackage{textcomp}
\usepackage{stfloats}
\usepackage{url}
\usepackage{verbatim}
\usepackage{svg}
\usepackage{amsmath}
\usepackage{import}
\usepackage{xcolor}

\usepackage{caption}
\usepackage{subcaption}
\usepackage{float}

\usepackage{booktabs}
\usepackage{tabularx}

\usepackage[numbers]{natbib}

\usepackage{lipsum} 
\usepackage{import}

\usepackage[switch]{lineno}

\newcommand{\paperID}{32}

\title{One-shot Detail Retouching with Patch Space Neural Transformation Blending}

\author{Fazilet~Gokbudak~and~Cengiz~Oztireli\\Department of Computer Science and Technology, University of Cambridge\\\{fg405, aco41\}@cam.ac.uk}

\begin{document}

\IEEEtitleabstractindextext{%
\begin{abstract}
Photo retouching is a difficult task for novice users as it requires expert knowledge and advanced tools. Photographers often spend a great deal of time generating high-quality retouched photos with intricate details. In this paper, we introduce a one-shot learning based technique to automatically retouch details of an input image based on just a single pair of before and after example images. Our approach provides accurate and generalizable detail edit transfer to new images. We achieve these by proposing a new representation for image to image maps. Specifically, we propose neural field based transformation blending in the patch space for defining patch to patch transformations for each frequency band. This parametrization of the map with anchor transformations and associated weights, and spatio-spectral localized patches, allows us to capture details well while staying generalizable. We evaluate our technique both on known ground truth filters and artist retouching edits. Our method accurately transfers complex detail retouching edits.
\end{abstract}

}

\maketitle


\begin{figure*}
     \includegraphics[width=\linewidth]{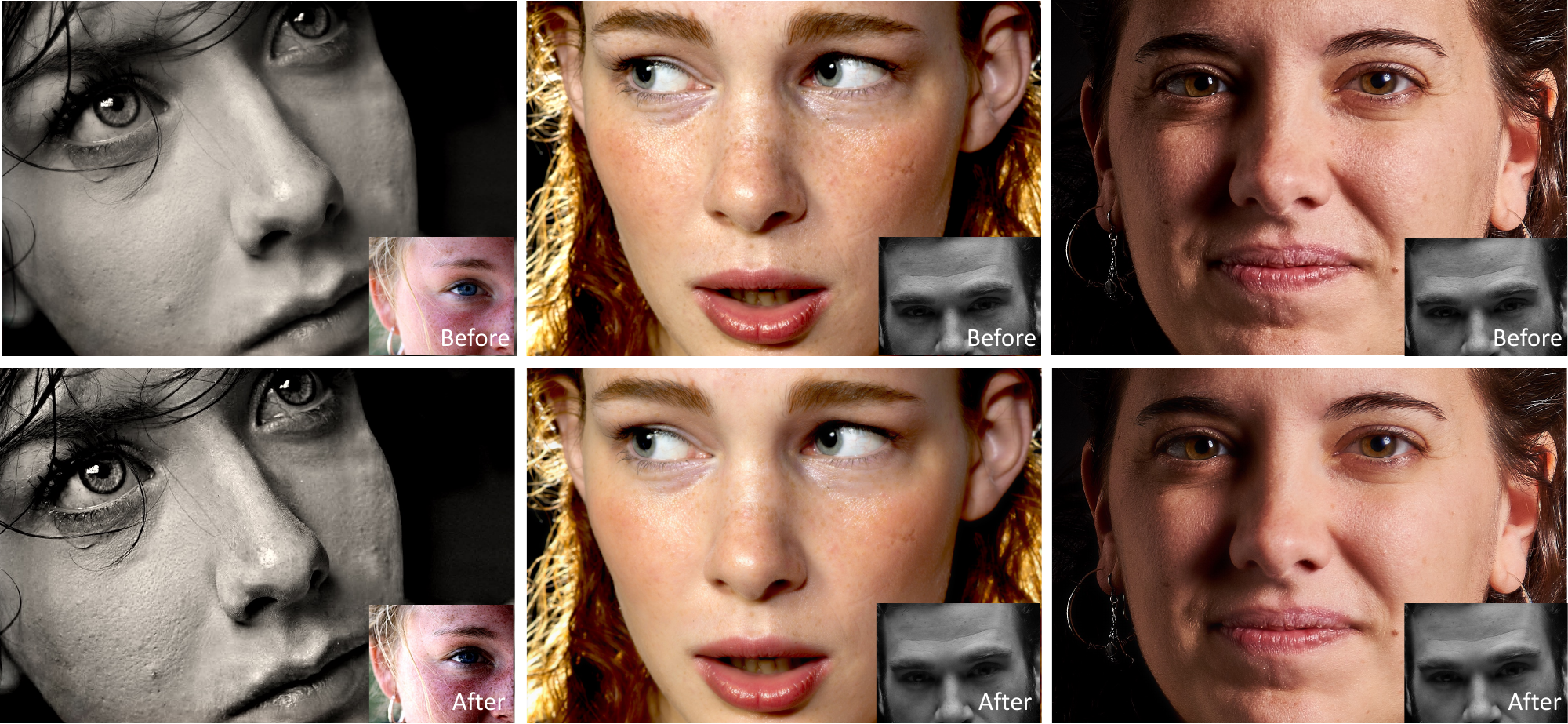}%
     \centering
      \caption{Our technique automatically transfers retouching edits to new images by learning the desired edits from one example \textit{before-after} pair (insets). The transferred edits accurately capture intricate details such as wrinkles, dark spots, strands of hair, or eyelashes, as shown in the input (top) and retouched (bottom) pairs.}
    \label{fig:teaser}
\end{figure*}

\IEEEraisesectionheading{
  \section{Introduction}\label{sec:introduction}
}

\IEEEPARstart{P}{hoto} retouching is often desirable as it improves the aesthetic quality of photographs by eliminating imperfections and highlighting subjects of interest. Even with significant progress in digital photography owing to advancements in camera sensors and image processing algorithms, professional retouches via manual adjustments are still needed to achieve a desired look. These artistic edits require considerable manual effort as they consist of global adjustments, such as brightening and contrast enhancement, as well as fine edits applied to local regions. Professionals spend a great deal of time to generate such edits, which motivates us to automatically mimic a specific style or type of retouch.

The development of automatic photo retouching tools can be helpful for both novice users and experts as it offers a basis for a professional retouching style. However, automating detailed edits of professionals is challenging as their editing pipelines are spatially varying, context-aware, and highly nonlinear, containing per-pixel adjustments. Recent learning-based methods address this complexity in image-to-image translation by proposing local context-aware methods, such as pixel-adaptive neural network architectures \cite{shaham2021spatially, li2020lapar}, learning parameters of local filters \cite{moran2020deeplpf}, or multi-stream models to extract global and local features separately \cite{Gharbi17Deep}. However, these data-driven methods require a large dataset of matching example image pairs. Even then, the mappings are sensitive to segmentation errors, unseen semantic regions, and image content~\cite{yan2016automatic}. 

Motivated by the gap between manual and automatic enhancement, we propose a novel photo retouching technique that can learn global and local adjustments from just \emph{a single example image pair}. Our method thus sidesteps the need for large datasets, which are very difficult to obtain for the detail retouching task. We allow users to choose one example \emph{before-after} pair from which our technique learns the underlying retouching style. Subsequently, we can apply the retouching edit to a different input image.

We assume that example and input images share similar local content. The user can thus decide on the semantics of the example and input photos and the structural changes to be transferred. This is easy for humans and practical for many scenarios, e.g. face edits transferred to faces. Our method then handles the difficult part for humans: capturing how fine details change in an edit and applying those automatically to a new image. The method can further be combined with brushes if fully automatic transfers are not desired. 

We achieve these by defining the retouching problem as a map that is given by a \emph{spatio-spectral patch-space neural field based transformation blending}. This representation is primarily inspired by professional detail retouching pipelines as we elaborate on in Section \ref{chap:motivations}. Our map representation is composed of learned patch maps at multiple scales, i.e. frequency bands. Each of these maps is represented by a number of \emph{transformation matrices} blended with \emph{patch-adaptive weights} that are represented as neural fields. We jointly optimize the transformation matrices and corresponding weights for each band. This representation captures edits to details better than any previous techniques while staying generalizable to new images. It is also simple enough to be extended in many different ways in future works.

In summary, there are two main contributions of this work: 

\begin{itemize}

    \item \textbf{A novel patch-space image map representation} as a blending of transformation matrices with neural fields.
    
	\item \textbf{A one-shot detail retouching algorithm} that allows transfer of edits to details to new images based on a single before-after image pair.

\end{itemize}

\section{Related Work}

Photo retouching has been explored in image processing and computer vision communities under different domains, such as photo enhancement and image-to-image translation. Below, we first discuss recent methods on photo enhancement and then image to image map definitions with the main focus on learning-based methods.

\subsection{Digital Photo Enhancement}

\textbf{Global image enhancement.} Color and tone transfer has been considered a very effective technique to improve the perceptual quality of photos with pre-defined rules or examples \cite{Faridul14ASurvey}. 
Earlier methods typically apply global changes and adjust image statistics \cite{Bychkovsky11Learning, Bae06Two, Pitie05NDimensional,Pitie07Automated,Reinhard01Color,Sunkavalli10Multi, he2020conditional, park2018distort}, e.g., mean and standard deviation, without considering image content and local variations \cite{CohenOr06Color}. These methods generally transfer color changes, ignoring edits in fine details. On the other hand, our method learns a mapping per frequency band, capturing transfers even in high frequencies. \citeauthor{Bychkovsky11Learning} \cite{Bychkovsky11Learning} collected the MIT-Adobe FiveK dataset of 5,000 photographs and their retouched versions by five artists. The authors propose a regression model to learn artists' retouching styles from before-retouched pairs. \citeauthor{chen2017fast} \cite{chen2017fast} introduce a fully-convolutional neural network model to learn global image processing operators, such as photographic style, nonlocal dehazing, and pencil drawing. In \cite{Hu18Exposure}, a photo retouching pipeline for various post-processing operations is presented, where global adjustment curves are approximated. The authors suggest a deep reinforcement learning approach to model users' edit preferences from a given photo collection. 
 
Nevertheless, global transfers cannot capture local and regional variations in a photo \cite{CohenOr06Color}.They may result in artifacts when the local target regions of the example and input images do not match. We adapt our mappings to each image patch separately, thus accurately capturing local edits in intricate details.

\textbf{Local context-aware image enhancement.} 
 To capture local variations, different methods have been proposed, such as learning local representative color transform \cite{kim2021representative}, estimating an image-to-illumination mapping with a local feature extractor \cite{wang2019underexposed}, local histogram matching \cite{Shapira13Image}, segmentation \cite{Laffont14Transient,Tai07Soft}, combining and learning pre-defined filters \cite{Berthouzoz11AFramework,Chen18Deep,Huang14Parametric,Omiya18Learning,Saeedi18Multimodal} or with further user guidance \cite{An10User,Pouli11Progressive,Tai05Local}, detection or learning of image semantics and context \cite{Gharbi17Deep,Hwang12Context,Kaufman12Content,Nam17Deep,Yan14Automatic,Zhu18Automatic}, matching \cite{HaCohen11Nonrigid}, or precise alignment \cite{Kagarlitsky09Piecewise, Shih13Data}. 
 
 Furthermore, recent work has focused on learning global and local adjustments via spatially-varying filters \cite{moran2020deeplpf, Gharbi17Deep, chen2018deep, shaham2021spatially, li2020lapar}. \citeauthor{chen2018deep} \cite{chen2018deep} introduce a global feature extraction layer along with per-pixel adjustments to enhance photos. Bilateral guided joint upsampling \cite{chen2016bilateral} also allows for local and global image processing with an encoder-decoder approach. HDRNet \cite{Gharbi17Deep} learn content-aware, global, and local adjustments via a two-stream convolutional architecture, which extracts local and global features separately to fit local affine transformations and encode the high-level description of images, respectively. Also, \citeauthor{moran2020deeplpf} \cite{moran2020deeplpf} propose to learn the parameters of three different spatially local filters to automatically enhance photos. 

Local color and tone adjustments might still be insufficient to capture intricate details \cite{Bae06Two}. Transfer of such details, in general, requires a dense matching \cite{HaCohen11Nonrigid} or alignment between example and input images \cite{Shih14Style}. To achieve either dense matching or alignment, methods constrain their datasets to contain very similar example and input images, for example, faces with similar characteristics and views~\cite{Shih14Style}. On the other hand, our method does not require dense correspondences between input and example images but still transfers intricate details. It accurately represents such complex mappings with an operator summing the effects of various transformations multiplied with corresponding patch-adaptive weights, applied at multiple frequency bands.

\textbf{Differentiable image processing pipelines.}
To have more flexibility and control over the rendering process, methods based on image signal processors (ISP)s have been proposed to enhance photos. In both \cite{tseng2019hyperparameter, yu2021reconfigisp}, hyperparameters of an ISP are optimized. Different from \cite{tseng2019hyperparameter}, which only applies to a fixed pipeline, \citeauthor{yu2021reconfigisp} \cite{yu2021reconfigisp} can explore different ISP architectures. Furthermore, \citeauthor{tseng2022neural} \cite{tseng2022neural} model a commercial raw processing pipeline with a series of neural networks to render sRGB images from raw inputs. As we assume example and input images to be processed RGB images rather than raw data, we refrain from comparing our method with such ISP-based methods.

\subsection{Defining Maps between Images}

\textbf{Unsupervised methods.} Some learning-based techniques only require one or more examples of retouched photos without their before examples to learn the transfer. Such unsupervised methods capture a certain style by decomposing images into a reflection map and an illumination map~\cite{ma2021retinexgan},
extracting and recomposing band representations of training images~\cite{yang2020fidelity}, regularizing  unpaired training using information extracted from the input~\cite{9334429}, segmenting the image into semantic regions~\cite{Liu16Makeup}, adaptive image regions~\cite{Frigo16Split}, learning semantic and global features~\cite{Chen18Deep}, progressively translating
image from coarse to fine via pyramids of generative models \cite{lin2020tuigan}, or utilizing artistic principles and pre-defined filters~\cite{Zhang13Style,Hu18Exposure}. These methods transfer pre-defined elements of the desired style, or global color and tone. Defining the desired style and the content of the input image that is to remain is challenging. Hence, these methods typically assume prior knowledge of the type of desired adjustments. Even then, capturing the retouching edits in details remains out of scope since these methods are typically designed for domain transfer, working on high level features of images. 

\textbf{Supervised methods.} 
For a conceivable representation, many supervised transfer methods require a large dataset of well-aligned example image pairs whose contents are very similar \cite{kim2021representative, wang2019underexposed}. However, finding or generating such a dataset is difficult as the content of images can change dramatically. Even with such a dataset, segmentation errors, unseen semantic regions, or image content can still change the results significantly \cite{10.1145/2790296}. In contrast, our method allows users to choose the example pairs from which the desired style is learned, hence sidestepping the challenging semantics problem. Similar content and structures between example and input images lead to more natural transfers.

Convolutional neural networks (CNNs) are the de-facto model for image processing with supervised learning methods. While CNNs present state-of-the-art results in computer vision tasks, they are not required \cite{tolstikhin2021mlp}. MLP-based architectures have recently gained popularity in image classification and image-to-image translation. \citeauthor{cazenavette2021mixergan} \cite{cazenavette2021mixergan} propose the MLP-Mixer architecture that only uses simple MLP blocks to learn image classification. \citeauthor{cazenavette2021mixergan} \cite{cazenavette2021mixergan} also show an application of an MLP-based architecture for image synthesis. They adapt the MLP-Mixer architecture \cite{tolstikhin2021mlp} to perform unpaired image-to-image translation. Our observation that MLP-based architectures attain competitive results in challenging vision tasks motivated us to explore the use of an MLP block as an alternative to CNNs in the context of photo retouching.

\section{Overview and Motivations}
\label{chap:motivations}

Given a pair of example images $X$ and $Y$, we aim to learn a map $M$ such that $Y = M(X)$. The learned map can then be applied to a new input image $I$ to obtain the retouched output $O = M(I)$. 

\begin{figure*}[th] 
	\includegraphics[width=\textwidth]{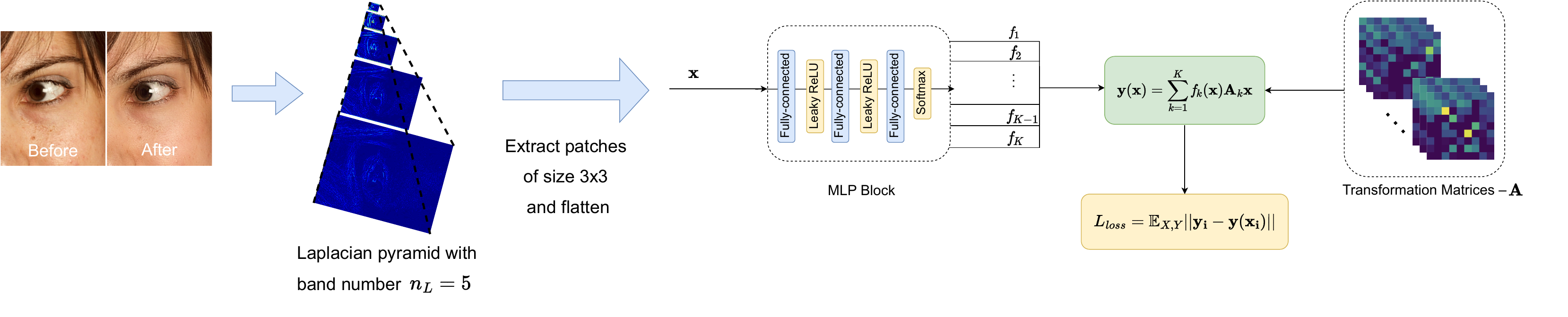}
    \caption{Our technique learns a separate mapping per frequency band by decomposing images into five different bands with a Laplacian pyramid. At each band $l$, we define a mapping between flattened patches $\mathbf{x}_i$, $\mathbf{y}_i$ extracted from before-after bands $X_l$, $Y_l$. Our field based method (MLP block) adapts transformations to input patches, providing local context-aware adjustments. All transformation matrices and MLP parameters are learned jointly from scratch per band for each before-after pair.}

\label{fig:modelT}
\end{figure*}

To define this map, we first decompose the example images into multiple feature maps $X_l$, $Y_l$ capturing details at different scales, such as coefficients at different bands of a Laplacian pyramid. We then define a separate mapping for each $X_l$, $Y_l$ pair in the patch space as a blending of transformation matrices with neural field based weights, all learned jointly. We illustrate the overall map representation in Figure~\ref{fig:modelT}.


For transfer of edits, the $M_l$ are computed and applied to each patch of the decomposition $I_l$ of an input image $I$ to obtain the corresponding output patch of $O_l$. The patches are finally placed at their spatial locations and averaged to reconstruct each image $O_l$ , which are summed to get the output image $O$.

Our motivation behind designing such a map representation with frequency decomposition and transformation blending comes from studying the nature of retouching edits. First, artists often decompose images into different frequency bands to have better control over structural and textural edits to details. Second, image patches of similar content, e.g. skin or hair, are retouched similarly. This means similar patches in the patch space translate into similar edits. Our representation leads to a different transformation for patches of differing content. Third, these edits are typically applied via brushes for smooth transitions. Our neural field based blending allows for such smooth interpolation, mimicking such brush strokes.

Although we are inspired by professional artist pipelines, we illustrate in the next sections that this new image to image mapping representation can replicate the effect of and transfer edits for many filters.




\section{One-shot Retouching}
\label{sec:Methodology}


\subsection{Frequency Decomposition}\label{sec:thePatchMap}

We first decompose example and input images into different frequency bands by constructing a Laplacian pyramid to capture details at multiple scales. In principle, it is possible to utilize any multiscale image decomposition method. However, we observed that a basic Laplacian pyramid helped us capture more accurate and generalizable results compared to a guided or bilateral pyramid. Therefore, we decompose images by

\begin{equation} 
	X_l = L_l(X) = 
 \left \{ \begin{aligned}
        X - G(2) \ast X \hspace{5mm} &l=0\\
        G(2^l) \ast X - G(2^{l+1}) \ast X \hspace{5mm} &l>0,
       \end{aligned}
 \right.
\end{equation}
where $G(\sigma)$ is the normalized Gaussian kernel, and $\ast$ denotes convolution. We also store the low-pass filtered image $S(X)$ such that $X = S(X) + \sum_{l=0}^{n_L} L_l(X)$. We then downsample each $L_l(X)$ and $S(X)$ according to the maximum frequency present at that band. This allows us to use small $3 \times 3$ patches at each band. In our experiments, we used $n_L = 5$ bands for the Laplacian pyramid.



Since each band is processed independently, we explain the steps of our technique below for two generic images $X$ and $Y$.



\subsection{Transformation Blending}\label{sec:Blending}

The mapping is defined between patches $\mathbf{x} \in \mathbb{R}^{d_X}$ to $\mathbf{y} \in \mathbb{R}^{d_Y}$ extracted from $X$ and $Y$, respectively, where we denote the patches with vectors stacking the pixel values and define the patch spaces as $\mathbb{R}^{d_X}$ and  $\mathbb{R}^{d_Y}$. For all results in this work, we work with $3 \times 3$ patches and thus $d_X = d_Y = 9$.

Our mapping takes the form of a weighted average of learned transformation matrices, where each transformation matrix is first multiplied with its corresponding blending weight: 



\begin{equation} 
	\mathbf{y} (\mathbf{x}) = \sum_{k=1}^K
	\mathit{f}_k (\mathbf{x}) \mathbf{A}_k \mathbf{x},
	\label{eq:weightedSum}
\end{equation} 
Here, $K$ is the number of transformation matrices, and $f_k$ are the blending weights, learned by an MLP block of output size $K$. The $\mathbf{A}_k$'s and $f_k$'s are jointly learned by minimizing the following loss on patches extracted from the before and after images.

\begin{equation}
    L_{loss}  = \mathbb{E}_{X, Y} || \mathbf{y_i} -   \mathbf{y} (\mathbf{x_i}) ||
\end{equation}

Each $\mathbf{A}_k$ corresponds to a different type of transformation and the $f_k(\mathbf{x})$'s, represented with the MLP, allow for a smooth transition between different transformations. The form of $f_k$'s is relatively simple with three fully-connected layers and nonlinear activation functions applied after each layer. This blending forms a simple but expressive transformation as we illustrate in the Section \ref{sec:results}.

\subsection{Retouching an Input Image}

We process the input image $I$ the same way as the before-after pair. First, we decompose the input into its Laplacian layers and then extract its patches per layer. After applying the learned mappings $M_l$ to the patches of the corresponding layers $L_{l}(I)$ independently, we then reconstruct the Laplacian bands of the output image $O_l$ by placing the patches at their spatial locations and averaging over the overlapping regions. Later, we obtain the final output image $O$ by summing the outputs $O_l$ and the residual of the input image:
\begin{equation}
    O = S(I) + \sum_{l=0}^{n_L} M_l(L_l(I))\,.
\end{equation}

\subsection{Implementation}
\label{sec:Implementation}
\begin{figure}[t!] 
    \centering
	\includegraphics[width=\columnwidth]{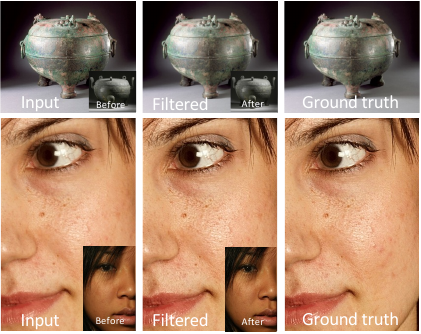}
    \caption{Our algorithm accurately represents simple algorithmic filters, such as Gaussian (top) and unsharp masking (bottom),  learned from a \textit{before-after} pair. The \textit{filtered} images obtained by applying the learned mapping to the \textit{input} images have very small difference to the ground truth images obtained by direct application of algorithmS.(PSNRs: 45.59 and 39.79 dB, respectively.)}

\label{fig:simpleAccuracy}
\end{figure}
\textbf{Patch size and stride.} In order to capture each frequency band at the right level of detail, we do not upsample the images $L_l(X)$ and use a small $3 \times 3$ patch size (with stride $1$). We experimented with larger patch sizes. However, this turned out to be counterproductive for the detail level we target, as details are blurred in larger patches. They also lead to overfitting and are harder to optimize for in general. We used a stride of $1$, and hence patches overlap on the image plane. The overlapping patches are averaged while reconstructing the image.

\begin{figure*}
  \centering
  \includegraphics[width=\linewidth]{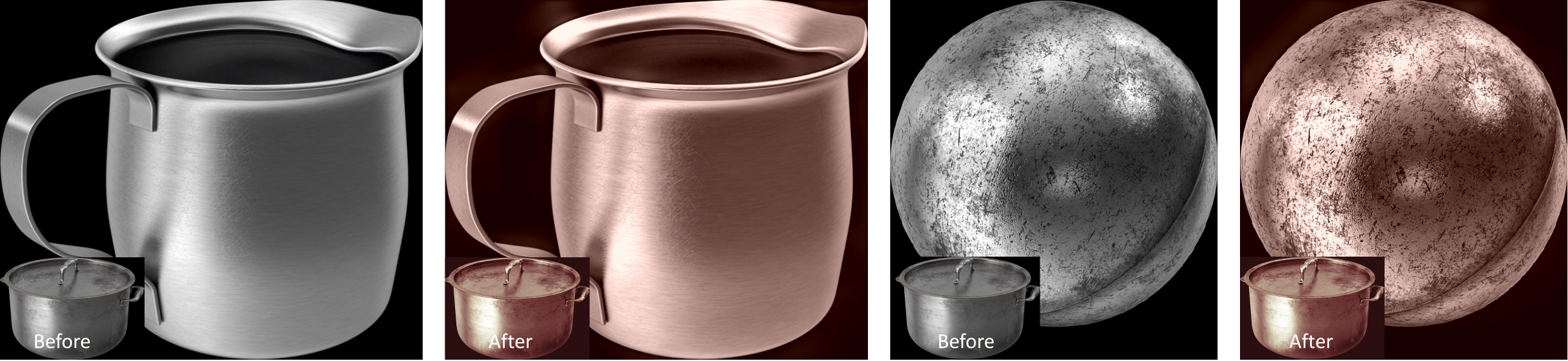}
  \caption{
           Material editing on photos (left), and rendered images (right), based on the examples provided (insets). Here, chrominance channels are also learned (see Section \ref{sec:Implementation}).}
           \label{fig:MaterialRes2}%
\end{figure*}

\textbf{Detail and color modifications.} We aim to capture intricate details present in highly detailed retouches and a wide range of image processing operators. Based on the observation that various operators can edit materials in the image space using the luminance component~\cite{Boyadzhiev15Band}, we focus on learning changes in luminance while preserving the input chrominance channels. In case desired retouching edits also involve color changes, we learn the mapping independently for luminance and chrominance channels to keep the dimensionality of the patch space low. We achieved satisfactory results with mild color changes (see Figure \ref{fig:MaterialRes2}).

\textbf{Evaluation metrics.} To quantitatively compare our method with state-of-the-art methods, we used PSNR and SSIM metrics. This is only possible if the before-after image pair was processed with a known, reproducible operator (see Section~\ref{sec:Comparisons} for details).


\textbf{Training details.}\label{train_det} We train different mappings with the same structure, defined in Equation \ref{eq:weightedSum}, for each frequency band of the Laplacian pyramid. Each mapping consists of one MLP block and $K$ number of transformation matrices, which are learned jointly per frequency band from scratch for each before-after pair. The MLP block employed in our experiments consists of three fully-connected layers and non-linearities applied after each layer. The output size of the last layer is the same as the number of transformation matrices.

To normalize the weights, we chose the last activation function to be Softmax, while for the first two layers, we apply Leaky ReLU. Each transformation matrix is randomly initialized with uniform distribution in the range $[0, 1]$. All experiments use the Adam optimizer with a learning rate of $10^{-2}$, which exponentially decays with a decay rate of $0.96$. We use $l_1$ loss function in all our experiments. 

\section{Results}
\label{sec:results}


\subsection{Ablation Study}\label{ablation}
The success of our learned mappings relies on two key components: patch-adaptive retouching and transformation blending. We thus conduct experiments to illustrate the significance of these.

\textbf{Transformation Matrices.} We compared transformation matrices of size $9 \times 9$ with scalar values. The method still remained spatially-varying, since we left the MLP the same, and used $K=256$ scalar weights. We tested both methods on 100 images and computed average PSNR values. We observed that our technique with matrices performed better than scalar values even in simple algorithmic filters, such as Gaussian and Unsharping Masking (around 2 dB and 3 dB higher PSNRs, respectively). 



\begin{figure}[th] 
    \centering
	\includegraphics[width=0.9\columnwidth]{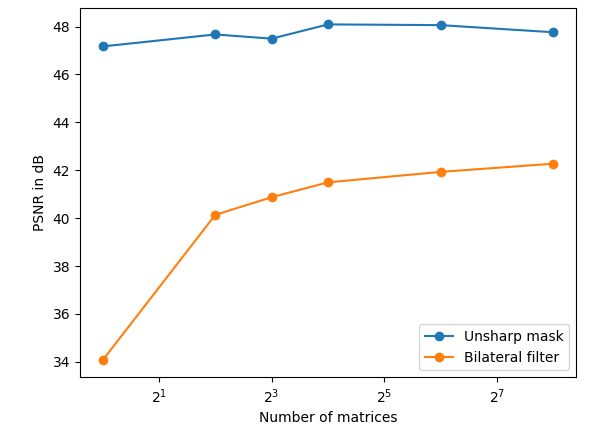}
    \caption{The higher the complexity of the learned algorithm, the more transformation matrices our technique requires to capture the effects on local regions accurately. While $K=1$ can be sufficient for our model to capture unsharp masking, it requires more matrices to represent bilateral filtering precisely.}
    \label{fig:ablation_K}
\end{figure}

As the complexity of a retouching style depends on multiple factors, such as artists’ design choices, user preferences, or the artist toolbox, it is challenging to analyze such effects on retouching examples quantitatively. For simple algorithmic filters, such as a Gaussian filter or unsharp masking, $K=1$ can sufficiently reproduce the filter. In contrast, more complex algorithms, such as a bilateral filter, require more matrices to capture the algorithmic edits accurately (Figure \ref{fig:ablation_K}). Since retouching edits combine the effect of multiple operators and are highly non-linear, we empirically chose $K=256$ for our retouching examples. 

\begin{figure}
\centering
\includegraphics[width=\columnwidth]{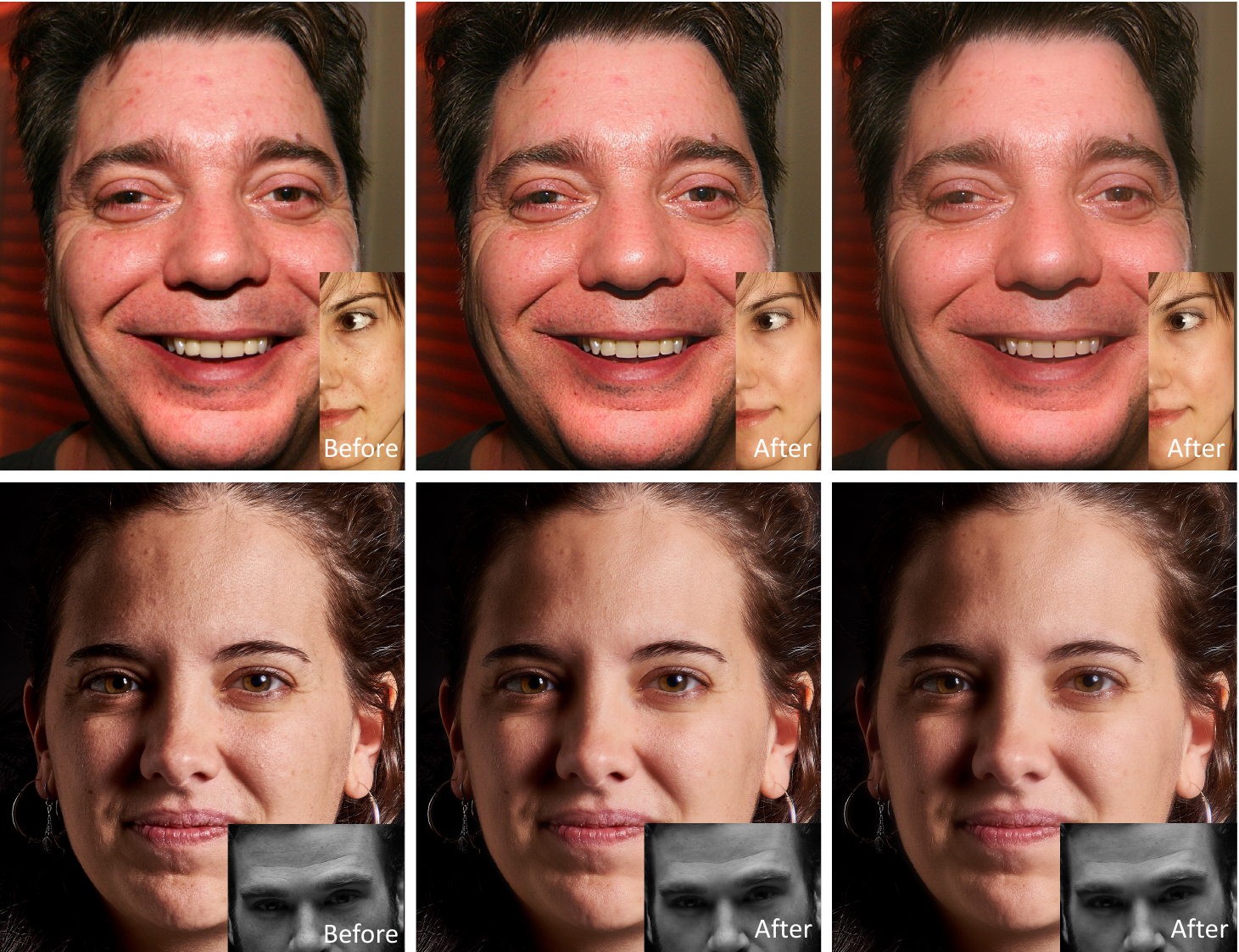}
    \caption{An MLP regressor cannot capture local edits, resulting in inaccurate retouching edits, such as blurring on the skin or around the eyes. Images in each row have the following order: input, filtered with our map design, filtered with an MLP regressor.}
\label{fig:ablation_MLP}
\end{figure}
\textbf{Patch-adaptive Transformation Blending.} We also compared our patch-adaptive mapping to an MLP regressor on the extracted patches. This directly learns the mapping from the decomposition of example before-after images instead of utilizing blended transformations. The MLP regressor follows a similar architecture as our MLP block (Figure \ref{fig:modelT}), with the only difference being the last activation function. We used Leaky ReLU here, since the Softmax function outputs pseudo-probabilities and is unsuitable for regression. Not explicitly handling the spatially-varying structure of the mapping and directly regressing limits the expressiveness of the model. This results in blurry results as shown in Figure~\ref{fig:ablation_MLP} because such a model cannot capture edits in intricate details, such as highlights around eyes and hair or brightening of the skin. We also tried increasing the capacity of the MLP regressor but did not observe much improvement in performance.




\subsection{Qualitative Results}
We tested our technique on a diverse range of before-after pairs, including face images from the \textbf{FFHQ} dataset \cite{DBLP:journals/corr/abs-1812-04948}. We focus on human portraits and face retouching in our experiments as they are arguably the most common and prioritized types of photos for retouching. We also illustrate that our technique provides visually pleasing results in different types of images, such as materials or rooms, and accurately captures image processing filters.

\begin{figure}[th] 
	\includegraphics[width=\columnwidth]{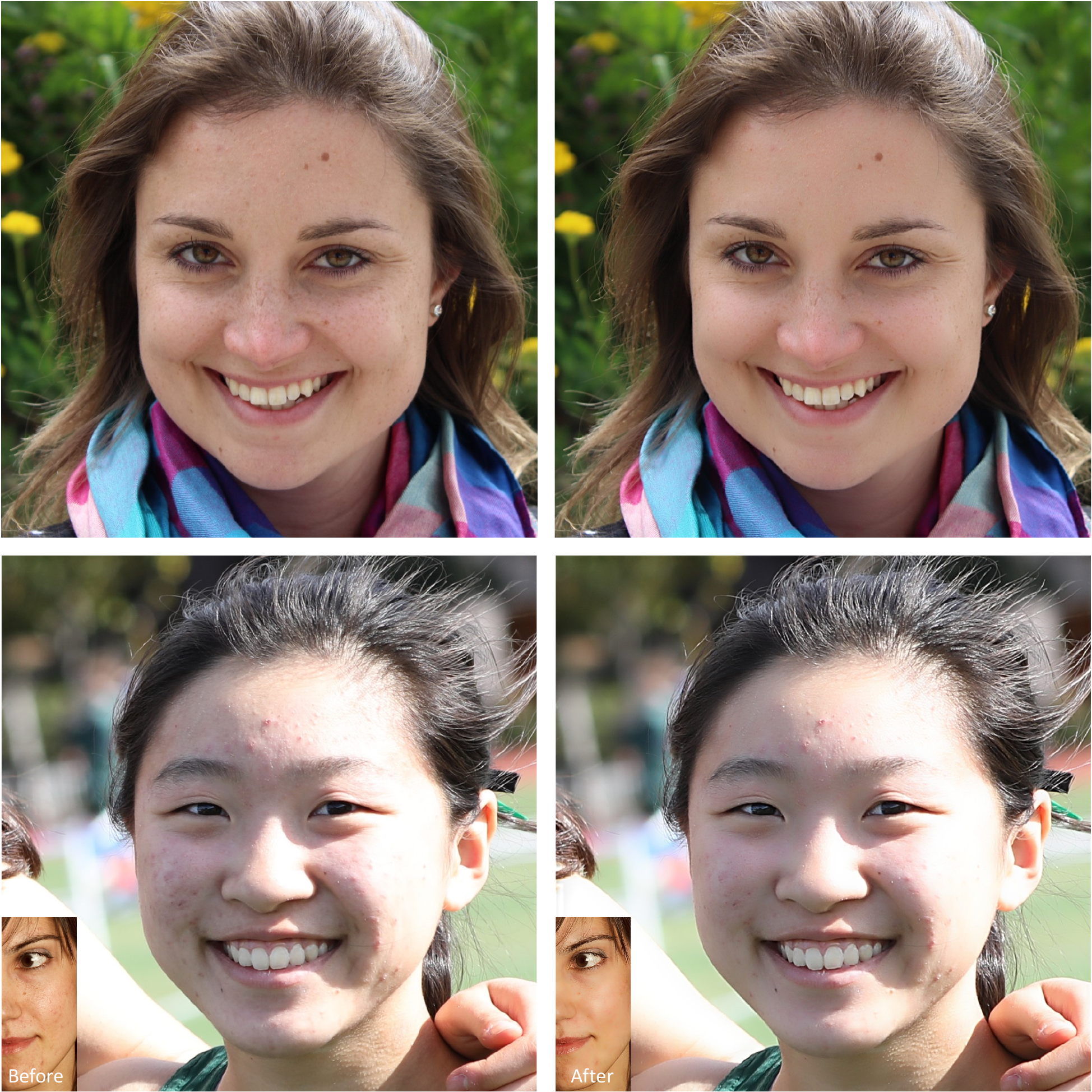}
    \caption{\label{fig:newdataset_ex}The reproduced retouching style from the example pair (inset) improves skin texture without affecting fine details, such as eyes and hair, for a visually improved portrait. Moreover, our technique generalizes well to faces with different lighting conditions and accurately reproduces the example retouching style.}
\end{figure}
\begin{figure}[th] 
	\includegraphics[width=\columnwidth]{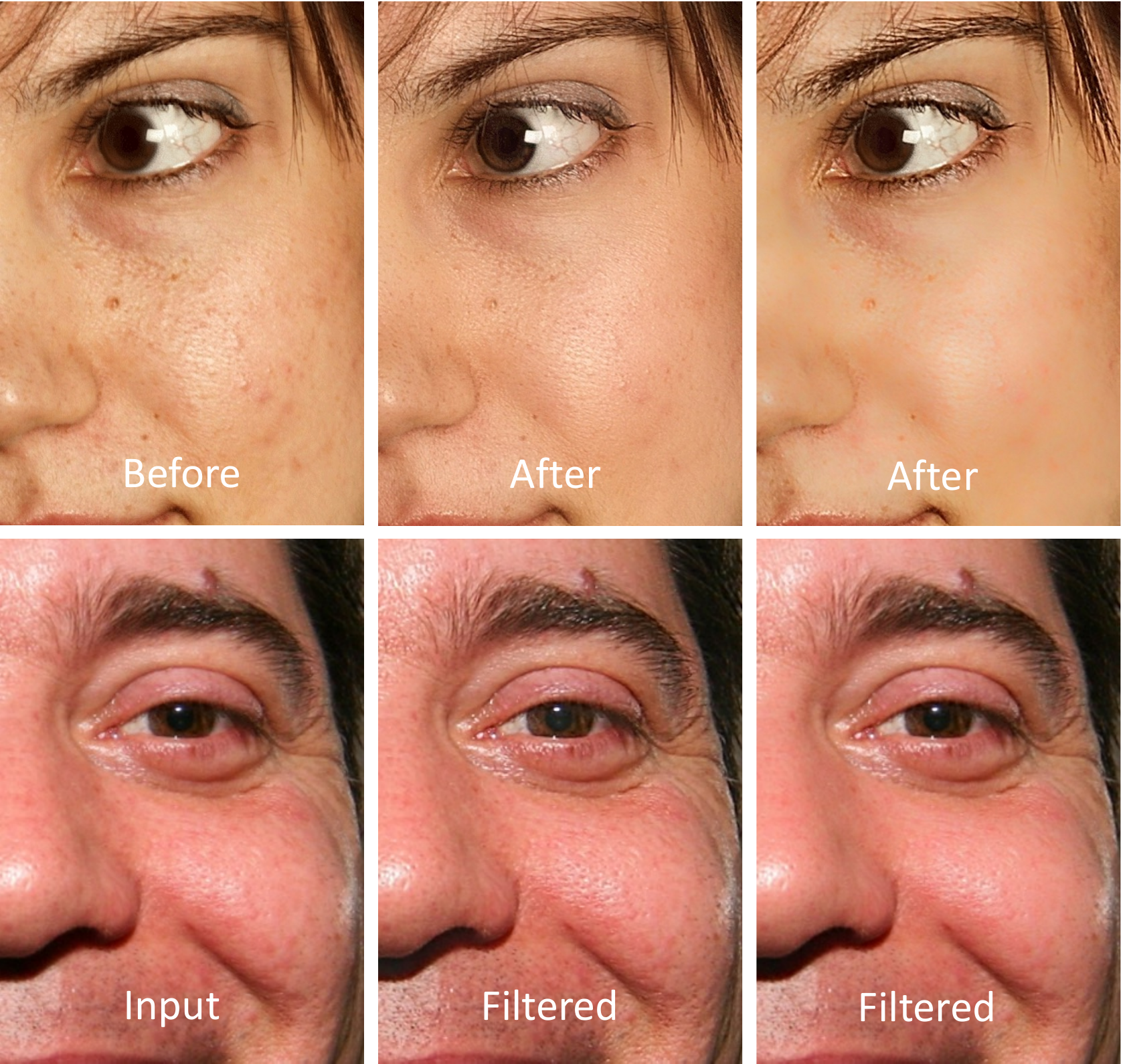}
    \caption{Our patch-adaptive technique can precisely reproduce nuances of retouching styles. The top and bottom rows show the example pairs and their corresponding retouched photos, respectively.}
\label{fig:retouchingstyles}
\end{figure}

Human faces pose a particular challenge for our technique. However, our model can still capture highly nonlinear retouching edits and generalizes well to different types of faces, view directions, and lighting conditions, as illustrated in Figures~\ref{fig:teaser}, ~\ref{fig:simpleAccuracy}, ~\ref{fig:newdataset_ex}, and ~\ref{fig:retouchingstyles}.


\begin{figure}[th] 
	\includegraphics[width=\columnwidth]{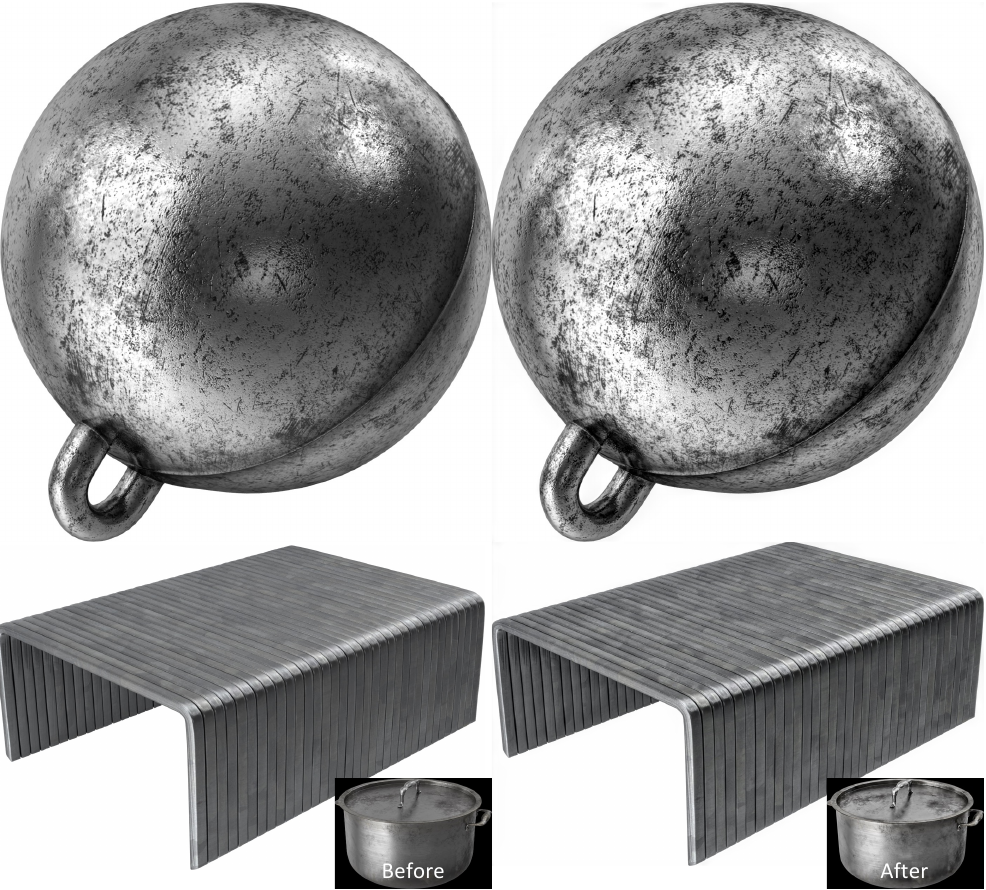}
    \caption{\label{fig:material_res}Material editing results on photos (left), and rendered images (right), based on the before-after pair (inset).}
\end{figure}

The example pairs in Figures~\ref{fig:teaser}, ~\ref{fig:newdataset_ex}  and ~\ref{fig:retouchingstyles} were generated by brushing onto the skin with artist created brushes, eye sharpening (sharpening example in Figure~\ref{fig:teaser}), and further brightness/contrast adjustments. These brushes first decompose the skin into a detail and base layer, typically with frequency decomposition, alter the detail layer and blend it with the base layer. They differ in how (1) they decompose the skin into the layers, i.e., what frequencies are in each layer, and (2) they edit and blend each layer with different opacity values. This variation creates retouching nuances, as shown in Figure~\ref{fig:retouchingstyles}. Our method can still accurately capture such slight differences in styles. 

In all our experiments, intricate details of the desired retouching, such as small-scale texture, eye, facial hair or material details, and global features, such as overall lighting and tone, are accurately reproduced. It is interesting to observe that the \emph{glamour} implied by, e.g., the example retouching in Figure~\ref{fig:newdataset_ex} is transferred from the example pair very accurately without causing an artificial look. Zooming into the skin reveals that pores and wrinkles are minimized, and the blemishes and discoloring of the skin are eliminated. At the same time, depending on the retouching edit, details, such as eyes or material texture, are more highlighted or preserved, and delicate features such as hair are preserved well (Figures ~\ref{fig:newdataset_ex}, ~\ref{fig:material_res} and ~\ref{fig:retouchingstyles}). 


In summary, our technique efficiently edits such intricate details, due to the significantly distinct local statistics of the texture at multiple scales, without affecting overlaying structures thanks to its spatially-varying nature and frequency decomposition. 




\subsection{Comparison with the state-of-the-art}
\label{sec:Comparisons}

Although there are various works related to automatic photo enhancement, to the best of our knowledge, none of them works with a single example pair for detail retouching. We thus compare our results with closely related automatic image-to-image translation methods, namely U-Net \cite{ronneberger2015u}, ASAPNet generator \cite{shaham2021spatially}, and Deep edge-aware filters \cite{xu2015deep}. 

We trained each network from scratch with one \textit{before-after} pair. To train the U-Net architecture, we changed the activation function of its last layer to ReLU and used $l_1$ loss function with Adam optimizer (same as ours). Similar to our method, ASAPNet is also a spatially-adaptive network. However, it is instead designed to hallucinate new details. Therefore, we similarly trained their generator model to ours with $l1$ loss, removing the discriminator. We observed that bilinear downsampling in their model causes checkerboard artifacts. Hence, we also removed this operator and learned an MLP per pixel, which caused the model to be highly complex with too many parameters. 


\newcommand\myworries[1]{\textcolor{red}{#1}}

\begin{table*}[th]
\centering
    \begin{tabular}{@{}cccccc@{}}
    \toprule

     \multicolumn{5}{c}{\textbf{Comparison results (PSNR in dB / SSIM)}} \\ \midrule
     \textbf{Filter Type} & \textbf{ASAPNet Generator} & \textbf{Deep Edge-aware} & \textbf{UNet} & \textbf{Ours}\\
     \midrule
     Gaussian & 39.42 / 0.985  &  36.11 / 0.962 & 40.60 / 0.981 & \textbf{41.36 / 0.986}\\
     Unsharp Mask & 29.79 / 0.891  & 30.23 / 0.902 & 32.01 / 0.923 & \textbf{33.81 / 0.939}\\
     Bilateral Filter & 33.79 / 0.938 & 33.09 / 0.935 & 34.15 / 0.941 & \textbf{39.23 / 0.971}\\
     Local Laplacian ($\alpha=2$, $\sigma =0.2$) & 30.91 / 0.915 & 30.29 /0.924 & 31.65 / 0.931 &  \textbf{33.93 / 0.955} \\
     Local Laplacian ($\alpha=0.5$, $\sigma =0.1$) & 31.87 / 0.908 & 30.91 / 0.892 & 32.95 / 0.926 & \textbf{35.79 / 0.932} 
     \\\bottomrule
    \end{tabular}
\caption{Quantitative performance comparison for the reproduction of various image processing filters. Average PSNR and SSIM values are computed over 182 images of different types of images including faces, landscapes, materials, and rooms.}
\label{tablecomparison}
\end{table*}

For a fair comparison with contemporary methods, we trained each network with the same example pair processed by four algorithmic filters: Gaussian, unsharp masking, Bilateral, and local Laplacian filters. As local Laplacian filters can perform a wide range of edge-aware operations, we apply two different versions of the filter, one for smoothing ($\alpha=2,  \sigma=0.2$) and one for enhancing details ($\alpha=0.5, \sigma=0.1$). Each network is trained from scratch with the same example pair resized to $256 \times 256$ for the corresponding filter. 
To prove the generalizability of our technique, we tested the models on different types of images, namely face images (100 images that are randomly sampled from MIT-Adobe FiveK \cite{Bychkovsky11Learning}), material images (22 images), room images (30), and landscape images (30). Each type was trained separately with its corresponding example pair. For instance, we trained an example pair of landscape images to test our model on landscape images. We evaluated the models using average PSNR and SSIM values. To generate the ground truths of the input images, we applied the same filter as applied to the before example image to obtain the after image. We trained each model in Y-channel after converting RGB images to their YCbCr versions and evaluated the results for Y-channel images. We duplicated the Y-channel in case the model requires three-channel images.

To obtain the UNet results for each type of images, we ran an additional experiment in which we changed the number of trainable parameters by removing some layers and trained the network from scratch for unsharp masking and bilateral filtering. The number of parameters we chose were 0.1M (with a few convolutional layers), 1.8M, 10M and 30M. For material images, we observed that 10M performed the best in terms of PSNR and SSIM values, while for other types of images 30M performed best. We tested the trained models on the images of the corresponding types and computed average PSNR values. Later, we chose the model with the best-performing parameters for each type of image for the quantitative comparison (Table \ref{tablecomparison}).

Overall, our method can outperform all architectures for each considered filter in terms of both PSNR and SSIM values. UNet shows the closest performance to our method, but their network capacity is significantly higher than ours (0.16M). Our technique proves more generalizable in learning different image processing operators from a single example pair with a lower model capacity.



\subsection{Limitations and Future Work}
\begin{figure}[th] 
\centering
	\includegraphics[width=\columnwidth]{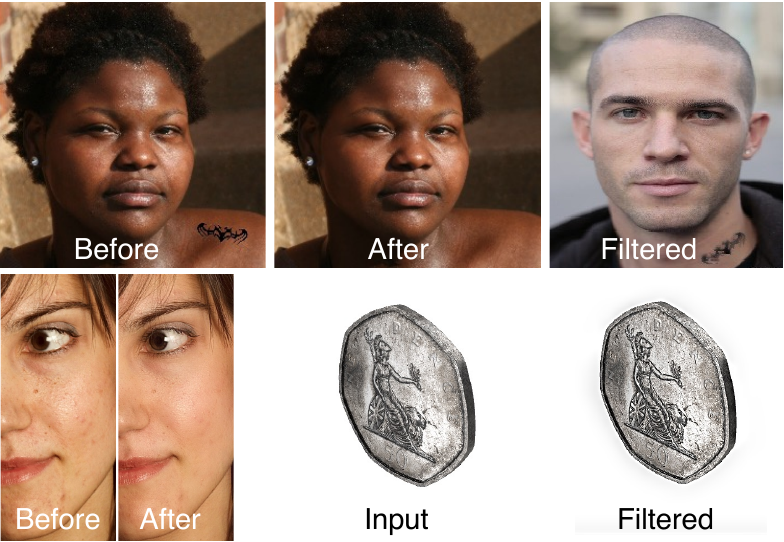}
    \caption{\label{fig:limitations}Our technique cannot accurately handle extreme non-repeating local effects such as tattoos (top), and when example and input images are of very different semantics (bottom).}
\end{figure}
A primary limitation of our work is its dependence on local patches at different scales, disregarding their spatial location. Hence, our method is most useful when details are retouched based on local and repeated characteristics of an image. Non-repeating spatially-dependent strong effects, e.g., tattoos or portrait stylizations with spatially varying lighting~\cite{Shih14Style}, cannot be handled by the current technique (see Figure \ref{fig:limitations}). We leave this as future work.


Since we rely on a single example image pair, transferring filters applied to arbitrary images~\cite{Yan14Automatic} is out of the scope of our current work. We require example and input images to have similar semantics for predictable transfer. Extending the technique to more than one pair of example images will require us to have consistently retouched details on all those example images. Finally, we require the example before and after images to be perfectly aligned. This requirement can be alleviated by incorporating an ICP~\cite{Besl92AMethod}-like approach into the optimization in Section~\ref{sec:Methodology}. 

Although our main focus in this paper is on artist-driven subjective retouching edits, the proposed technique is general. It can be applied to summarize and transfer arbitrary image transformations, significantly where details are modified. We are thus planning to investigate our technique further as a general transfer method for image-to-image translation. The patch-adaptive nature of our mappings makes them amenable to analysis.

\section{Conclusions}
We presented a neural field based technique for example-based automatic retouching of images. By formulating the transfer problem in the patch space, we showed that blending multiple transformation matrices with patch-adaptive weights can be utilized to learn an accurate and generalizable map. This allowed us to use images of different scenes, people, views, and environmental conditions as the example pair and input. We illustrated the technique's utility on various retouching examples. We believe that our image map representation can be helpful in many other image processing tasks.

\begin{figure*}
  \centering
  \includegraphics[height=0.96\textheight, width=\linewidth,keepaspectratio]{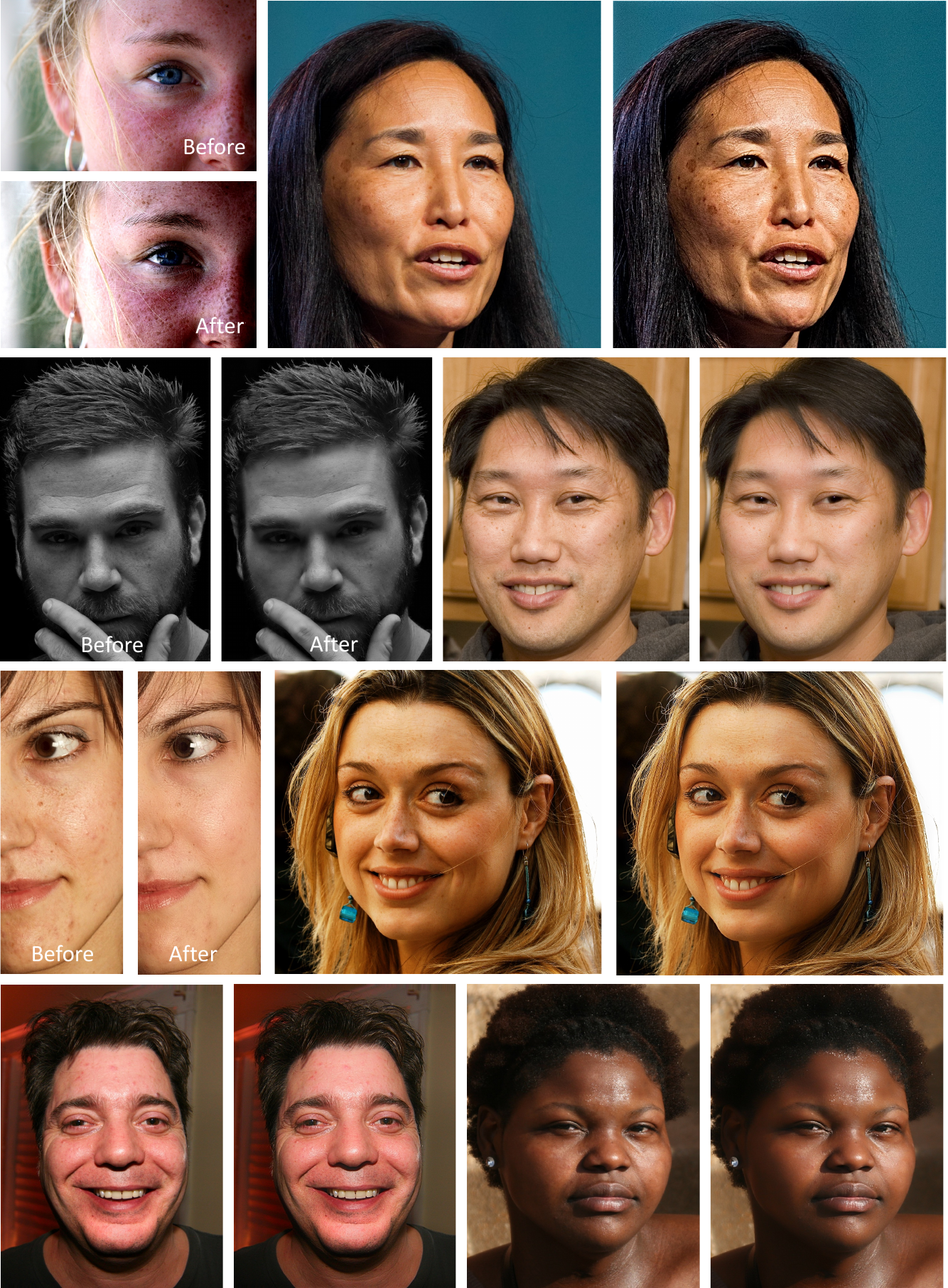}
  \caption{
           Retouches reproduced by our algorithm based on single before-after pairs. }
           \label{fig:AdditionalRes}%
\end{figure*}

\begin{figure*}
  \centering
  \includegraphics[width=\linewidth]{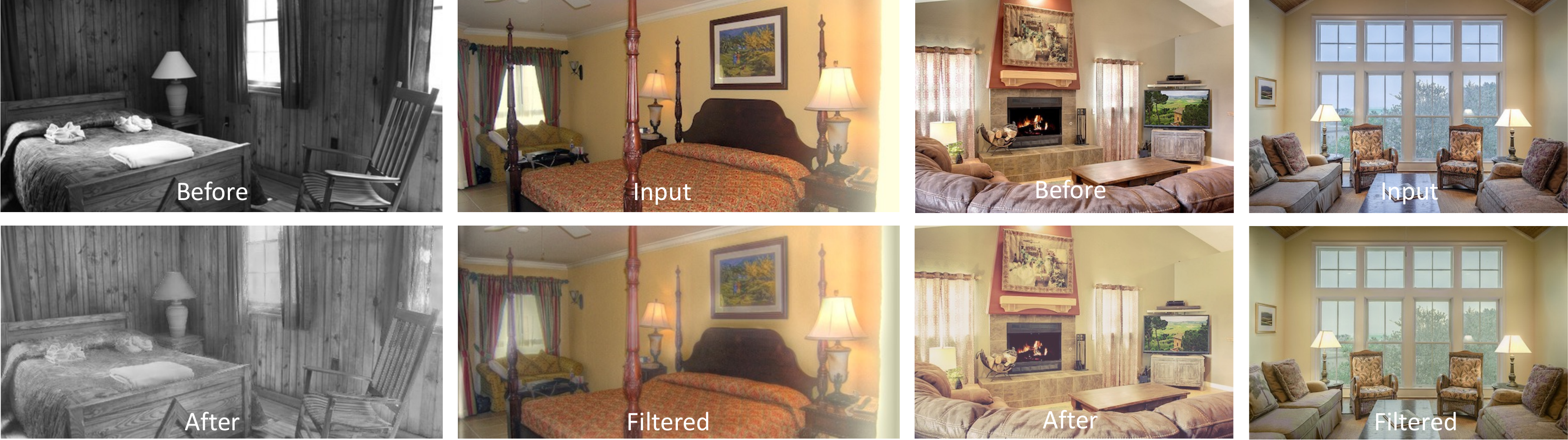}
    \caption{
           Our technique can consistently capture filters applied to examples of various scenes.}
   \label{fig:Roomfilter}%
\end{figure*}

\ifpeerreview \else
\section*{Acknowledgments}
We thank Dr. Mauricio Delbracio for useful discussions. We also thank Param Hanji for feedback on the manuscript.
This work was supported by a UKRI Future Leaders Fellowship [grant number G104084].
\fi

\bibliographystyle{IEEEtranN}
\bibliography{main}

\begin{thebibliography}{59}
\providecommand{\natexlab}[1]{#1}
\providecommand{\url}[1]{#1}
\csname url@samestyle\endcsname
\providecommand{\newblock}{\relax}
\providecommand{\bibinfo}[2]{#2}
\providecommand{\BIBentrySTDinterwordspacing}{\spaceskip=0pt\relax}
\providecommand{\BIBentryALTinterwordstretchfactor}{4}
\providecommand{\BIBentryALTinterwordspacing}{\spaceskip=\fontdimen2\font plus
\BIBentryALTinterwordstretchfactor\fontdimen3\font minus
  \fontdimen4\font\relax}
\providecommand{\BIBforeignlanguage}[2]{{%
\expandafter\ifx\csname l@#1\endcsname\relax
\typeout{** WARNING: IEEEtranN.bst: No hyphenation pattern has been}%
\typeout{** loaded for the language `#1'. Using the pattern for}%
\typeout{** the default language instead.}%
\else
\language=\csname l@#1\endcsname
\fi
#2}}
\providecommand{\BIBdecl}{\relax}
\BIBdecl

\bibitem[Shaham et~al.(2021)Shaham, Gharbi, Zhang, Shechtman, and
  Michaeli]{shaham2021spatially}
T.~R. Shaham, M.~Gharbi, R.~Zhang, E.~Shechtman, and T.~Michaeli,
  ``Spatially-adaptive pixelwise networks for fast image translation,'' in
  \emph{Proceedings of the IEEE/CVF Conference on Computer Vision and Pattern
  Recognition}, 2021, pp. 14\,882--14\,891.

\bibitem[Li et~al.(2020)Li, Zhou, Qi, Jiang, Lu, and Jia]{li2020lapar}
W.~Li, K.~Zhou, L.~Qi, N.~Jiang, J.~Lu, and J.~Jia, ``Lapar: Linearly-assembled
  pixel-adaptive regression network for single image super-resolution and
  beyond,'' \emph{Advances in Neural Information Processing Systems}, vol.~33,
  pp. 20\,343--20\,355, 2020.

\bibitem[Moran et~al.(2020)Moran, Marza, McDonagh, Parisot, and
  Slabaugh]{moran2020deeplpf}
S.~Moran, P.~Marza, S.~McDonagh, S.~Parisot, and G.~Slabaugh, ``Deeplpf: Deep
  local parametric filters for image enhancement,'' in \emph{Proceedings of the
  IEEE/CVF Conference on Computer Vision and Pattern Recognition}, 2020, pp.
  12\,826--12\,835.

\bibitem[Gharbi et~al.(2017)Gharbi, Chen, Barron, Hasinoff, and
  Durand]{Gharbi17Deep}
M.~Gharbi, J.~Chen, J.~T. Barron, S.~W. Hasinoff, and F.~Durand, ``Deep
  bilateral learning for real-time image enhancement,'' \emph{ACM Trans.
  Graph.}, vol.~36, no.~4, pp. 118:1--118:12, Jul. 2017.

\bibitem[Yan et~al.(2016{\natexlab{a}})Yan, Zhang, Wang, Paris, and
  Yu]{yan2016automatic}
Z.~Yan, H.~Zhang, B.~Wang, S.~Paris, and Y.~Yu, ``Automatic photo adjustment
  using deep neural networks,'' \emph{ACM Transactions on Graphics (TOG)},
  vol.~35, no.~2, pp. 1--15, 2016.

\bibitem[Faridul et~al.(2014)Faridul, Pouli, Chamaret, Stauder, Tremeau, and
  Reinhard]{Faridul14ASurvey}
H.~S. Faridul, T.~Pouli, C.~Chamaret, J.~Stauder, A.~Tremeau, and E.~Reinhard,
  ``{A Survey of Color Mapping and its Applications},'' in \emph{Eurographics
  2014 - State of the Art Reports}, S.~Lefebvre and M.~Spagnuolo, Eds.\hskip
  1em plus 0.5em minus 0.4em\relax The Eurographics Association, 2014.

\bibitem[Bychkovsky et~al.(2011)Bychkovsky, Paris, Chan, and
  Durand]{Bychkovsky11Learning}
V.~Bychkovsky, S.~Paris, E.~Chan, and F.~Durand, ``Learning photographic global
  tonal adjustment with a database of input/output image pairs,'' in
  \emph{Proceedings of the 2011 IEEE Conference on Computer Vision and Pattern
  Recognition}, ser. CVPR '11.\hskip 1em plus 0.5em minus 0.4em\relax
  Washington, DC, USA: IEEE Computer Society, 2011, pp. 97--104.

\bibitem[Bae et~al.(2006)Bae, Paris, and Durand]{Bae06Two}
S.~Bae, S.~Paris, and F.~Durand, ``Two-scale tone management for photographic
  look,'' \emph{ACM Trans. Graph.}, vol.~25, no.~3, pp. 637--645, Jul. 2006.

\bibitem[Pitie et~al.(2005)Pitie, Kokaram, and Dahyot]{Pitie05NDimensional}
F.~Pitie, A.~Kokaram, and R.~Dahyot, ``N-dimensional probability density
  function transfer and its application to color transfer,'' in \emph{Computer
  Vision, 2005. ICCV 2005. Tenth IEEE International Conference on}, vol.~2,
  Oct. 2005, pp. 1434--1439 Vol. 2.

\bibitem[Piti{\'e} et~al.(2007)Piti{\'e}, Kokaram, and
  Dahyot]{Pitie07Automated}
F.~Piti{\'e}, A.~C. Kokaram, and R.~Dahyot, ``Automated colour grading using
  colour distribution transfer,'' \emph{Comput. Vis. Image Underst.}, vol. 107,
  no. 1-2, pp. 123--137, Jul. 2007.

\bibitem[Reinhard et~al.(2001)Reinhard, Adhikhmin, Gooch, and
  Shirley]{Reinhard01Color}
E.~Reinhard, M.~Adhikhmin, B.~Gooch, and P.~Shirley, ``Color transfer between
  images,'' \emph{Computer Graphics and Applications, IEEE}, vol.~21, no.~5,
  pp. 34--41, Sep. 2001.

\bibitem[Sunkavalli et~al.(2010)Sunkavalli, Johnson, Matusik, and
  Pfister]{Sunkavalli10Multi}
K.~Sunkavalli, M.~K. Johnson, W.~Matusik, and H.~Pfister, ``Multi-scale image
  harmonization,'' \emph{ACM Trans. Graph.}, vol.~29, no.~4, pp. 125:1--125:10,
  Jul. 2010.

\bibitem[He et~al.(2020)He, Liu, Qiao, and Dong]{he2020conditional}
J.~He, Y.~Liu, Y.~Qiao, and C.~Dong, ``Conditional sequential modulation for
  efficient global image retouching,'' in \emph{European Conference on Computer
  Vision}.\hskip 1em plus 0.5em minus 0.4em\relax Springer, 2020, pp. 679--695.

\bibitem[Park et~al.(2018)Park, Lee, Yoo, and Kweon]{park2018distort}
J.~Park, J.-Y. Lee, D.~Yoo, and I.~S. Kweon, ``Distort-and-recover: Color
  enhancement using deep reinforcement learning,'' in \emph{Proceedings of the
  IEEE conference on computer vision and pattern recognition}, 2018, pp.
  5928--5936.

\bibitem[Cohen-Or et~al.(2006)Cohen-Or, Sorkine, Gal, Leyvand, and
  Xu]{CohenOr06Color}
D.~Cohen-Or, O.~Sorkine, R.~Gal, T.~Leyvand, and Y.-Q. Xu, ``Color
  harmonization,'' \emph{ACM Trans. Graph.}, vol.~25, no.~3, pp. 624--630, Jul.
  2006.

\bibitem[Chen et~al.(2017)Chen, Xu, and Koltun]{chen2017fast}
Q.~Chen, J.~Xu, and V.~Koltun, ``Fast image processing with fully-convolutional
  networks,'' in \emph{Proceedings of the IEEE International Conference on
  Computer Vision}, 2017, pp. 2497--2506.

\bibitem[Hu et~al.(2018)Hu, He, Xu, Wang, and Lin]{Hu18Exposure}
\BIBentryALTinterwordspacing
Y.~Hu, H.~He, C.~Xu, B.~Wang, and S.~Lin, ``Exposure: A white-box photo
  post-processing framework,'' \emph{ACM Trans. Graph.}, vol.~37, no.~2, pp.
  26:1--26:17, May 2018. [Online]. Available:
  \url{http://doi.acm.org/10.1145/3181974}
\BIBentrySTDinterwordspacing

\bibitem[Kim et~al.(2021)Kim, Choi, Kim, and Koh]{kim2021representative}
H.~Kim, S.-M. Choi, C.-S. Kim, and Y.~J. Koh, ``Representative color transform
  for image enhancement,'' in \emph{Proceedings of the IEEE/CVF International
  Conference on Computer Vision}, 2021, pp. 4459--4468.

\bibitem[Wang et~al.(2019)Wang, Zhang, Fu, Shen, Zheng, and
  Jia]{wang2019underexposed}
R.~Wang, Q.~Zhang, C.-W. Fu, X.~Shen, W.-S. Zheng, and J.~Jia, ``Underexposed
  photo enhancement using deep illumination estimation,'' in \emph{Proceedings
  of the IEEE/CVF Conference on Computer Vision and Pattern Recognition}, 2019,
  pp. 6849--6857.

\bibitem[Shapira et~al.(2013)Shapira, Avidan, and Hel-Or]{Shapira13Image}
D.~Shapira, S.~Avidan, and Y.~Hel-Or, ``Multiple histogram matching,'' in
  \emph{Image Processing (ICIP), 2013 20th IEEE International Conference on},
  Sep. 2013, pp. 2269--2273.

\bibitem[Laffont et~al.(2014)Laffont, Ren, Tao, Qian, and
  Hays]{Laffont14Transient}
P.-Y. Laffont, Z.~Ren, X.~Tao, C.~Qian, and J.~Hays, ``Transient attributes for
  high-level understanding and editing of outdoor scenes,'' \emph{ACM Trans.
  Graph.}, vol.~33, no.~4, pp. 149:1--149:11, Jul. 2014.

\bibitem[Tai et~al.(2007)Tai, Jia, and Tang]{Tai07Soft}
Y.~W. Tai, J.~Jia, and C.~K. Tang, ``Soft color segmentation and its
  applications,'' \emph{IEEE Transactions on Pattern Analysis and Machine
  Intelligence}, vol.~29, no.~9, pp. 1520--1537, Sep. 2007.

\bibitem[Berthouzoz et~al.(2011)Berthouzoz, Li, Dontcheva, and
  Agrawala]{Berthouzoz11AFramework}
\BIBentryALTinterwordspacing
F.~Berthouzoz, W.~Li, M.~Dontcheva, and M.~Agrawala, ``A framework for
  content-adaptive photo manipulation macros: Application to face, landscape,
  and global manipulations,'' \emph{ACM Trans. Graph.}, vol.~30, no.~5, pp.
  120:1--120:14, Oct. 2011. [Online]. Available:
  \url{http://doi.acm.org/10.1145/2019627.2019639}
\BIBentrySTDinterwordspacing

\bibitem[Chen et~al.(2018{\natexlab{a}})Chen, Wang, Kao, and
  Chuang]{Chen18Deep}
Y.-S. Chen, Y.-C. Wang, M.-H. Kao, and Y.-Y. Chuang, ``Deep photo enhancer:
  Unpaired learning for image enhancement from photographs with gans,'' in
  \emph{The IEEE Conference on Computer Vision and Pattern Recognition (CVPR)},
  Jun. 2018.

\bibitem[Huang et~al.(2014)Huang, Zhang, Lai, Kopf, Cohen-Or, and
  Hu]{Huang14Parametric}
S.-S. Huang, G.-X. Zhang, Y.-K. Lai, J.~Kopf, D.~Cohen-Or, and S.-M. Hu,
  ``Parametric meta-filter modeling from a single example pair,'' \emph{Vis.
  Comput.}, vol.~30, no. 6-8, pp. 673--684, Jun. 2014.

\bibitem[Omiya et~al.(2018)Omiya, Simo-Serra, Iizuka, and
  Ishikawa]{Omiya18Learning}
M.~Omiya, E.~Simo-Serra, S.~Iizuka, and H.~Ishikawa, ``{Learning Photo
  Enhancement by Black-Box Model Optimization Data Generation},'' in
  \emph{SIGGRAPH Asia 2018 Technical Briefs}, 2018.

\bibitem[Saeedi et~al.(2018)Saeedi, Hoffman, DiVerdi, Ghandeharioun, Johnson,
  and Adams]{Saeedi18Multimodal}
\BIBentryALTinterwordspacing
A.~Saeedi, M.~D. Hoffman, S.~J. DiVerdi, A.~Ghandeharioun, M.~J. Johnson, and
  R.~P. Adams, ``Multimodal prediction and personalization of photo edits with
  deep generative models,'' in \emph{Proceedings of the 21st International
  Conference on Artificial Intelligence and Statistics (AISTATS)}, 2018,
  arXiv:1704.04997 [stat.ML]. [Online]. Available:
  \url{http://www.cs.princeton.edu/~rpa/pubs/saeedi2018multimodal.pdf}
\BIBentrySTDinterwordspacing

\bibitem[An and Pellacini(2010)]{An10User}
X.~An and F.~Pellacini, ``User-controllable color transfer,'' \emph{Computer
  Graphics Forum}, vol.~29, no.~2, pp. 263--271, 2010.

\bibitem[Pouli and Reinhard(2011)]{Pouli11Progressive}
T.~Pouli and E.~Reinhard, ``Progressive color transfer for images of arbitrary
  dynamic range,'' \emph{Computers \& Graphics}, vol.~35, no.~1, pp. 67 -- 80,
  2011, extended Papers from Non-Photorealistic Animation and Rendering (NPAR)
  2010.

\bibitem[Tai et~al.(2005)Tai, Jia, and Tang]{Tai05Local}
Y.-W. Tai, J.~Jia, and C.-K. Tang, ``Local color transfer via probabilistic
  segmentation by expectation-maximization,'' in \emph{Computer Vision and
  Pattern Recognition, 2005. CVPR 2005. IEEE Computer Society Conference on},
  vol.~1, Jun. 2005, pp. 747--754 vol. 1.

\bibitem[Hwang et~al.(2012)Hwang, Kapoor, and Kang]{Hwang12Context}
S.~J. Hwang, A.~Kapoor, and S.~B. Kang, ``Context-based automatic local image
  enhancement,'' in \emph{Proceedings of the 12th European Conference on
  Computer Vision - Volume Part I}, ser. ECCV 2012.\hskip 1em plus 0.5em minus
  0.4em\relax Berlin, Heidelberg: Springer-Verlag, 2012, pp. 569--582.

\bibitem[Kaufman et~al.(2012)Kaufman, Lischinski, and Werman]{Kaufman12Content}
L.~Kaufman, D.~Lischinski, and M.~Werman, ``Content-aware automatic photo
  enhancement,'' \emph{Comput. Graph. Forum}, vol.~31, no.~8, pp. 2528--2540,
  Dec. 2012.

\bibitem[Nam and Kim(2017)]{Nam17Deep}
S.~Nam and S.~J. Kim, ``Deep semantics-aware photo adjustment,'' \emph{CoRR},
  vol. abs/1706.08260, 2017.

\bibitem[Yan et~al.(2014)Yan, Zhang, Wang, Paris, and Yu]{Yan14Automatic}
Z.~Yan, H.~Zhang, B.~Wang, S.~Paris, and Y.~Yu, ``Automatic photo adjustment
  using deep learning,'' \emph{CoRR}, vol. abs/1412.7725, 2014.

\bibitem[Zhu and Yu(2018)]{Zhu18Automatic}
F.~Zhu and Y.~Yu, ``Automatic image stylization using deep fully convolutional
  networks,'' \emph{CoRR}, vol. abs/1811.10872, 2018.

\bibitem[HaCohen et~al.(2011)HaCohen, Shechtman, Goldman, and
  Lischinski]{HaCohen11Nonrigid}
Y.~HaCohen, E.~Shechtman, D.~B. Goldman, and D.~Lischinski, ``Non-rigid dense
  correspondence with applications for image enhancement,'' \emph{ACM Trans.
  Graph.}, vol.~30, no.~4, pp. 70:1--70:10, Jul. 2011.

\bibitem[Kagarlitsky et~al.(2009)Kagarlitsky, Moses, and
  Hel-Or]{Kagarlitsky09Piecewise}
S.~Kagarlitsky, Y.~Moses, and Y.~Hel-Or, ``Piecewise-consistent color mappings
  of images acquired under various conditions,'' in \emph{Computer Vision, 2009
  IEEE 12th International Conference on}, Sep. 2009, pp. 2311--2318.

\bibitem[Shih et~al.(2013)Shih, Paris, Durand, and Freeman]{Shih13Data}
Y.~Shih, S.~Paris, F.~Durand, and W.~T. Freeman, ``Data-driven hallucination of
  different times of day from a single outdoor photo,'' \emph{ACM Trans.
  Graph.}, vol.~32, no.~6, pp. 200:1--200:11, Nov. 2013.

\bibitem[Chen et~al.(2018{\natexlab{b}})Chen, Wang, Kao, and
  Chuang]{chen2018deep}
Y.-S. Chen, Y.-C. Wang, M.-H. Kao, and Y.-Y. Chuang, ``Deep photo enhancer:
  Unpaired learning for image enhancement from photographs with gans,'' in
  \emph{Proceedings of the IEEE Conference on Computer Vision and Pattern
  Recognition}, 2018, pp. 6306--6314.

\bibitem[Chen et~al.(2016)Chen, Adams, Wadhwa, and Hasinoff]{chen2016bilateral}
J.~Chen, A.~Adams, N.~Wadhwa, and S.~W. Hasinoff, ``Bilateral guided
  upsampling,'' \emph{ACM Transactions on Graphics (TOG)}, vol.~35, no.~6, pp.
  1--8, 2016.

\bibitem[Shih et~al.(2014)Shih, Paris, Barnes, Freeman, and
  Durand]{Shih14Style}
Y.~Shih, S.~Paris, C.~Barnes, W.~T. Freeman, and F.~Durand, ``Style transfer
  for headshot portraits,'' \emph{ACM Trans. Graph.}, vol.~33, no.~4, pp.
  148:1--148:14, Jul. 2014.

\bibitem[Tseng et~al.(2019)Tseng, Yu, Yang, Mannan, Arnaud, Nowrouzezahrai,
  Lalonde, and Heide]{tseng2019hyperparameter}
E.~Tseng, F.~Yu, Y.~Yang, F.~Mannan, K.~S. Arnaud, D.~Nowrouzezahrai, J.-F.
  Lalonde, and F.~Heide, ``Hyperparameter optimization in black-box image
  processing using differentiable proxies.'' \emph{ACM Trans. Graph.}, vol.~38,
  no.~4, pp. 27--1, 2019.

\bibitem[Yu et~al.(2021)Yu, Li, Peng, Loy, and Gu]{yu2021reconfigisp}
K.~Yu, Z.~Li, Y.~Peng, C.~C. Loy, and J.~Gu, ``Reconfigisp: Reconfigurable
  camera image processing pipeline,'' in \emph{Proceedings of the IEEE/CVF
  International Conference on Computer Vision}, 2021, pp. 4248--4257.

\bibitem[Tseng et~al.(2022)Tseng, Zhang, Jebe, Zhang, Xia, Fan, Heide, and
  Chen]{tseng2022neural}
E.~Tseng, Y.~Zhang, L.~Jebe, X.~Zhang, Z.~Xia, Y.~Fan, F.~Heide, and J.~Chen,
  ``Neural photo-finishing,'' \emph{ACM Transactions on Graphics (TOG)},
  vol.~41, no.~6, pp. 1--15, 2022.

\bibitem[Ma et~al.(2021)Ma, Guo, Yu, Chen, Ren, Xi, Li, and
  Zhou]{ma2021retinexgan}
T.~Ma, M.~Guo, Z.~Yu, Y.~Chen, X.~Ren, R.~Xi, Y.~Li, and X.~Zhou, ``Retinexgan:
  Unsupervised low-light enhancement with two-layer convolutional decomposition
  networks,'' \emph{IEEE Access}, vol.~9, pp. 56\,539--56\,550, 2021.

\bibitem[Yang et~al.(2020)Yang, Wang, Fang, Wang, and Liu]{yang2020fidelity}
W.~Yang, S.~Wang, Y.~Fang, Y.~Wang, and J.~Liu, ``From fidelity to perceptual
  quality: A semi-supervised approach for low-light image enhancement,'' in
  \emph{Proceedings of the IEEE/CVF conference on computer vision and pattern
  recognition}, 2020, pp. 3063--3072.

\bibitem[Jiang et~al.(2021)Jiang, Gong, Liu, Cheng, Fang, Shen, Yang, Zhou, and
  Wang]{9334429}
Y.~Jiang, X.~Gong, D.~Liu, Y.~Cheng, C.~Fang, X.~Shen, J.~Yang, P.~Zhou, and
  Z.~Wang, ``Enlightengan: Deep light enhancement without paired supervision,''
  \emph{IEEE Transactions on Image Processing}, vol.~30, pp. 2340--2349, 2021.

\bibitem[{Liu} et~al.(2016){Liu}, {Ou}, {Qian}, {Wang}, and {Cao}]{Liu16Makeup}
S.~{Liu}, X.~{Ou}, R.~{Qian}, W.~{Wang}, and X.~{Cao}, ``{Makeup like a
  superstar: Deep Localized Makeup Transfer Network},'' \emph{ArXiv e-prints},
  Apr. 2016.

\bibitem[Frigo et~al.(2016)Frigo, Sabater, Delon, and Hellier]{Frigo16Split}
O.~Frigo, N.~Sabater, J.~Delon, and P.~Hellier, ``{Split and Match:
  Example-based Adaptive Patch Sampling for Unsupervised Style Transfer},''
  Mar. 2016, peer-reviewed paper accepted to be presented at IEEE International
  Conference on Computer Vision and Pattern Recognition (CVPR), Jun 2016, Las
  Vegas, United States.

\bibitem[Lin et~al.(2020)Lin, Pang, Xia, Chen, and Luo]{lin2020tuigan}
J.~Lin, Y.~Pang, Y.~Xia, Z.~Chen, and J.~Luo, ``Tuigan: Learning versatile
  image-to-image translation with two unpaired images,'' in \emph{Computer
  Vision--ECCV 2020: 16th European Conference, Glasgow, UK, August 23--28,
  2020, Proceedings, Part IV 16}.\hskip 1em plus 0.5em minus 0.4em\relax
  Springer, 2020, pp. 18--35.

\bibitem[Zhang et~al.(2013)Zhang, Cao, Chen, Liu, and Tang]{Zhang13Style}
W.~Zhang, C.~Cao, S.~Chen, J.~Liu, and X.~Tang, ``Style transfer via image
  component analysis,'' \emph{IEEE Transactions on Multimedia}, vol.~15, no.~7,
  pp. 1594--1601, Nov. 2013.

\bibitem[Yan et~al.(2016{\natexlab{b}})Yan, Zhang, Wang, Paris, and
  Yu]{10.1145/2790296}
\BIBentryALTinterwordspacing
Z.~Yan, H.~Zhang, B.~Wang, S.~Paris, and Y.~Yu, ``Automatic photo adjustment
  using deep neural networks,'' \emph{ACM Trans. Graph.}, vol.~35, no.~2, Feb.
  2016. [Online]. Available: \url{https://doi.org/10.1145/2790296}
\BIBentrySTDinterwordspacing

\bibitem[Tolstikhin et~al.(2021)Tolstikhin, Houlsby, Kolesnikov, Beyer, Zhai,
  Unterthiner, Yung, Steiner, Keysers, Uszkoreit, et~al.]{tolstikhin2021mlp}
I.~O. Tolstikhin, N.~Houlsby, A.~Kolesnikov, L.~Beyer, X.~Zhai, T.~Unterthiner,
  J.~Yung, A.~Steiner, D.~Keysers, J.~Uszkoreit \emph{et~al.}, ``Mlp-mixer: An
  all-mlp architecture for vision,'' \emph{Advances in Neural Information
  Processing Systems}, vol.~34, pp. 24\,261--24\,272, 2021.

\bibitem[Cazenavette and De~Guevara(2021)]{cazenavette2021mixergan}
G.~Cazenavette and M.~L. De~Guevara, ``Mixergan: An mlp-based architecture for
  unpaired image-to-image translation,'' \emph{arXiv preprint
  arXiv:2105.14110}, 2021.

\bibitem[Boyadzhiev et~al.(2015)Boyadzhiev, Bala, Paris, and
  Adelson]{Boyadzhiev15Band}
I.~Boyadzhiev, K.~Bala, S.~Paris, and E.~Adelson, ``Band-sifting decomposition
  for image-based material editing,'' \emph{ACM Trans. Graph.}, vol.~34, no.~5,
  pp. 163:1--163:16, Nov. 2015.

\bibitem[Karras et~al.(2018)Karras, Laine, and
  Aila]{DBLP:journals/corr/abs-1812-04948}
\BIBentryALTinterwordspacing
T.~Karras, S.~Laine, and T.~Aila, ``A style-based generator architecture for
  generative adversarial networks,'' \emph{CoRR}, vol. abs/1812.04948, 2018.
  [Online]. Available: \url{http://arxiv.org/abs/1812.04948}
\BIBentrySTDinterwordspacing

\bibitem[Ronneberger et~al.(2015)Ronneberger, Fischer, and
  Brox]{ronneberger2015u}
O.~Ronneberger, P.~Fischer, and T.~Brox, ``U-net: Convolutional networks for
  biomedical image segmentation,'' in \emph{Medical Image Computing and
  Computer-Assisted Intervention--MICCAI 2015: 18th International Conference,
  Munich, Germany, October 5-9, 2015, Proceedings, Part III 18}.\hskip 1em plus
  0.5em minus 0.4em\relax Springer, 2015, pp. 234--241.

\bibitem[Xu et~al.(2015)Xu, Ren, Yan, Liao, and Jia]{xu2015deep}
L.~Xu, J.~Ren, Q.~Yan, R.~Liao, and J.~Jia, ``Deep edge-aware filters,'' in
  \emph{International Conference on Machine Learning}.\hskip 1em plus 0.5em
  minus 0.4em\relax PMLR, 2015, pp. 1669--1678.

\bibitem[Besl and McKay(1992)]{Besl92AMethod}
P.~J. Besl and N.~D. McKay, ``A method for registration of 3-d shapes,''
  \emph{IEEE Trans. Pattern Anal. Mach. Intell.}, vol.~14, no.~2, pp. 239--256,
  Feb. 1992.

\end{thebibliography}




\end{document}